\documentclass[runningheads]{llncs}
\usepackage[T1]{fontenc}
\usepackage{pifont}
\usepackage{booktabs}
\newcommand{\etal}{\emph{et al.}}
\usepackage{subcaption}
\usepackage{caption}
\usepackage{booktabs}
\usepackage{amsmath}
\usepackage{amssymb}
\usepackage{hyperref} 
\usepackage{graphicx}
\begin{document}
\title{Reading in the Dark: \\Low-light Scene Text Recognition}
%
%
\author{Xuanshuo Fu \and
 Lei Kang\and
Ernest Valveny\and \\
Dimosthenis Karatzas\and
Javier Vazquez-Corral}
\authorrunning{X. Fu et al.}
%
\institute{{Computer Vision Center, Barcelona, Spain\\Universitat Autònoma de Barcelona, Barcelona, Spain\\}
\email{\{xuanshuo, lkang, ernest, dimos, javier.vazquez\}@cvc.uab.es}}
\maketitle              
\begin{abstract}
Accurate text recognition in low-light environments is essential for intelligent systems in applications ranging from autonomous vehicles to smart surveillance. However, challenges such as poor illumination and noise interference remain underexplored. To address this gap, we introduce \textbf{LSTR}, a large-scale Low-light Scene Text Recognition dataset comprising 11,273 low-light images generated from well-lit datasets (ICDAR2015, IIIT5K, and WordArt), along with \textbf{ESTR}, which includes 60 real nighttime street-scene images in English and Spanish for exclusive evaluation. We explore two solution strategies: (1) employing Optical Character Recognition (OCR) models with fine-tuning and LoRA-based fine-tuning and (2) a joint training strategy that integrates a low-light image enhancement (LLIE) module with an OCR model. In particular, we propose a novel re-render LLIE (\textbf{RLLIE}) module, which demonstrates improved performance on real-world data. Through extensive experimentation, we analyze various training strategies and address a key research question: \emph{How bright is bright enough for effective scene text recognition?} Our results indicate that standalone LLIE or OCR models perform inadequately under low-light conditions, highlighting the advantages of specialized, jointly trained text-centric approaches. Additionally, we provide a comprehensive benchmark to support future research in robust low-light scene text recognition. \url{https://huggingface.co/datasets/lumimusta/Low-light_Scene_Text_Dataset}.

\end{abstract}
\section{Introduction}
\begin{figure}
  \centering
  \includegraphics[width=0.89\linewidth]{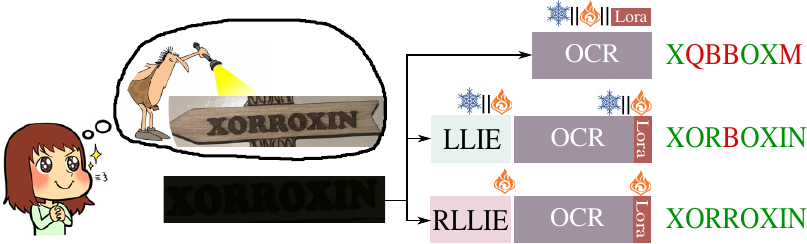}
  \caption{For low-light scene text images, torchlight can make text readable, but real-world conditions lack sufficient illuminance. Two main approaches address this: using only an OCR model, or enhancing the image to increase brightness. We propose a novel end-to-end method that integrates our Re-render LLIE (RLLIE) with an OCR module.}
  \vspace{-5mm}
  \label{fig:arch}
\end{figure}

Scene text recognition in low-light conditions is a challenging computer vision problem. Images captured at night or in poorly lit environments often suffer from low contrast, noise, and color distortions, which severely degrade the performance of standard optical character recognition (OCR) models. While generic low-light image enhancement algorithms can brighten dark images, they are typically designed to improve human visual perception rather than text legibility. Consequently, directly applying off-the-shelf enhancement methods as a pre-processing step may inadvertently weaken or distort critical text features, leading to suboptimal recognition results~\cite{xu2024seeing}. Existing scene text recognizers trained on well-lit data also struggle in low-light scenarios, as the domain gap causes missed or mispredicted characters. There is a clear need for specialized approaches to recognize scene text \emph{in the dark}, going beyond naive brighten-and-recognize pipelines.

A straightforward strategy for low-light scene text recognition is to cascade a low-light image enhancement (LLIE) algorithm with a text recognizer. However, this two-stage approach has notable drawbacks. Generic LLIE methods (e.g. histogram equalization or Retinex-based enhancers) are not optimized for text, often introducing artifacts or over-smoothing that hamper subsequent text recognition. Recently developed learning-based LLIE models such as Zero-DCE~\cite{guo2020zero}, Zero-DCE++~\cite{li2021learning}, RUAS~\cite{liu2021retinex}, SCI~\cite{ma2022toward}, or Generative diffusion prior (GDP)~\cite{fei2023generative} achieve impressive brightness correction on natural images without references. Yet, they might fail to preserve fine details like character strokes, as their loss functions focus on visual quality rather than text readability. More task-specific attempts have emerged in the context of text detection under low light, Xue~\etal~\cite{xue2020arbitrarily} developed a method to enhance image visibility by fusing spatial and frequency-domain features and then detect text via MSER, but this classical pipeline struggles with complex scenes. Hsu~\etal~\cite{hsu2022extremely} showed that state-of-the-art enhancers (RetinexNet~\cite{wei2018deep}, CycleGAN~\cite{zhu2017unpaired}, etc.) often cannot restore text regions sufficiently for detectors, and proposed an enhancement network with attention and edge modules plus a text-focused loss to better recover characters in extremely dark images. However, even with such tailored enhancement, a separate OCR stage might not fully exploit the potential of joint optimization. End-to-end trainable solutions are needed to bridge the gap between enhancement and recognition for low-light text.

In this paper, we propose an end-to-end architecture for low-light scene text recognition that combines an image enhancer and a text recognizer into one unified model. Specifically, we design a Re-Render Low-Light Image Enhancement (RLLIE) module that learns to produce a brightness-adjusted version of the input image, which is fed into an OCR recognition network. Unlike a fixed pre-processor, the RLLIE module is jointly trained with the OCR, allowing the enhancement to be guided by the text recognition loss. This synergy ensures that the enhanced image is not only visually brighter but also optimally suited for text extraction. We investigate a key research question in this context: \textbf{How bright is bright enough for effective scene text recognition?} Our findings indicate that brighter images do not necessarily lead to improved recognition performance, highlighting the need for task-oriented optimization rather than simply maximizing image brightness.

To evaluate our approach, we construct two datasets: a large synthetic low-light dataset \textbf{LSTR} and a manually labeled dataset \textbf{ESTR}. The ~\textbf{ESTR} dataset comprises real nighttime scene images, covering various text instances with different fonts, languages (English and Spanish), and backgrounds under diverse low-light conditions.

In summary, our contributions are as follows:

\begin{enumerate}
    \item We systematically define the new task by simulating low-light images from standard datasets (ICDAR2015 \cite{peyrard2015icdar2015}, IIIT5K \cite{mishra2012scene}, and WordArt \cite{xie2022toward}), resulting in an 11,273-image LLSTR dataset. Additionally, we provide an evaluation set consisting of 60 real low-light images captured in low-light cordition.
    \item We explore two groups of solutions: (a) OCR model-based methods, including using pre-trained OCR models, fine-tuning, and LoRA fine-tuning; (b) OCR approaches integrated with low-light image enhancement (LLIE), encompassing reconstruction of normal-light images using a pre-trained LLIE model, joint training of LLIE and OCR models with mutual updates, and our novel re-render LLIE (RLLIE) module trained jointly with an OCR model.
    \item We perform comprehensive theoretical and experimental analyses to assess LLSTR, thereby offering valuable benchmarks for the research community. 
\end{enumerate}

\section{Related Work}

\subsection{Low-Light Image Enhancement}
Low-light image enhancement (LLIE) has been extensively studied, yielding a variety of approaches. Early methods include histogram equalization and gamma correction techniques, as well as model-based strategies derived from Retinex theory~\cite{liu2021retinex}. While effective for general contrast improvement, these classical methods often produce noise or inconsistent enhancement when applied to complex scenes. In recent years, learning-based methods have advanced LLIE significantly. Supervised methods trained on paired dark/bright images (e.g., EnlightenGAN~\cite{jiang2021enlightengan}, KinD~\cite{zhang2019kindling}, RetinexNet~\cite{wei2018deep}, R2RNet~\cite{hai2023r2rnet}) can produce impressive visual results, but require laborious data collection and may not generalize to unobserved conditions. This has spurred interest in zero-reference or unsupervised methods that rely only on unpaired or no reference images.

Notably, Guo~\etal~\cite{guo2020zero} proposed Zero-DCE, which enhances low-light images by learning pixel-wise adjustment curves under non-reference losses, avoiding the need for paired ground truth. Li~\etal~\cite{li2021learning} further introduced Zero-DCE++ with a tiny network and improved training for faster inference. In a different direction, Retinex-inspired methods have also shown strong performance: RUAS~\cite{liu2021retinex} adopts an unrolling framework with neural architecture search to obtain a lightweight enhancer, and SCI~\cite{ma2022toward} employs cascaded illumination learning with weight sharing for efficient enhancement.

More recently, diffusion models have been explored for LLIE, e.g., GDP~\cite{fei2023generative} leverages a pre-trained diffusion prior for zero-shot enhancement. In addition, SRDACE~\cite{wen2023self} combines Retinex-style illumination modeling with learnable curve adjustment to improve brightness and contrast. While these methods achieve strong general enhancement, they may still fail to consistently preserve text-specific details (e.g., fine strokes and sharp edges) that are critical for scene text recognition, motivating task-aware designs discussed next.

\subsection{Scene Text Detection and Recognition in Low Light}

There is growing interest in scene text detection under low-light conditions, which is closely related to our recognition setting. Xue~\etal~\cite{xue2020arbitrarily} explored low-light text detection by coupling spatial--frequency enhancement with a classical MSER-based detector, but the reliance on hand-crafted components limits robustness in cluttered backgrounds. Recent deep methods instead seek illumination-invariant representations within the detection pipeline. LIST (Low Illumination Scene Text detector)~\cite{liu2022list} synthesizes low-light training samples via a non-reference augmentation network and adopts a dual-path CNN to process the original and an intensified version of the image, with learned feature fusion; together with the introduced DarkText dataset, it substantially improves detection robustness in dark scenes. In contrast, Seeing Text in the Dark~\cite{xu2024seeing} avoids explicit pre-enhancement and regularizes detector training with auxiliary constraints to preserve textual cues during down-sampling, yielding representations more resilient to low-light degradations.

Beyond detection, several works target enhancement tailored for text readability. Hsu~\etal~\cite{hsu2022extremely} proposed an enhancer for extremely dark images with text restoration, using attention/edge cues and a detector-guided loss obtained by backpropagating through a fixed text detector (CRAFT). Lin~\etal~\cite{lin2025text} extends this direction with edge-aware attention and supervised curve-based low-light synthesis for augmentation. Collectively, these studies show that incorporating text-specific objectives into enhancement—via multi-task learning or specialized losses—can better align visual improvement with OCR performance~\cite{nguyen2024diffusion}. Diffusion-based approaches such as DiD achieve strong results but are computationally demanding and require careful guidance tuning to avoid over-smoothing characters or retaining noise. Different from DiD, we adopt a direct end-to-end discriminative formulation: a deterministic enhancer is trained jointly with the OCR loss, enabling simpler and more efficient inference. We nevertheless include DiD as a strong baseline and empirically demonstrate that our integrated RLLIE--OCR strategy achieves favorable accuracy--efficiency trade-offs.

\section{Methodology}


\begin{figure*}[t]
\centering

\begin{minipage}[t]{0.41\textwidth}
  \vspace{0pt}\centering

  \begin{subfigure}{\linewidth}
    \centering
    \includegraphics[width=\linewidth]{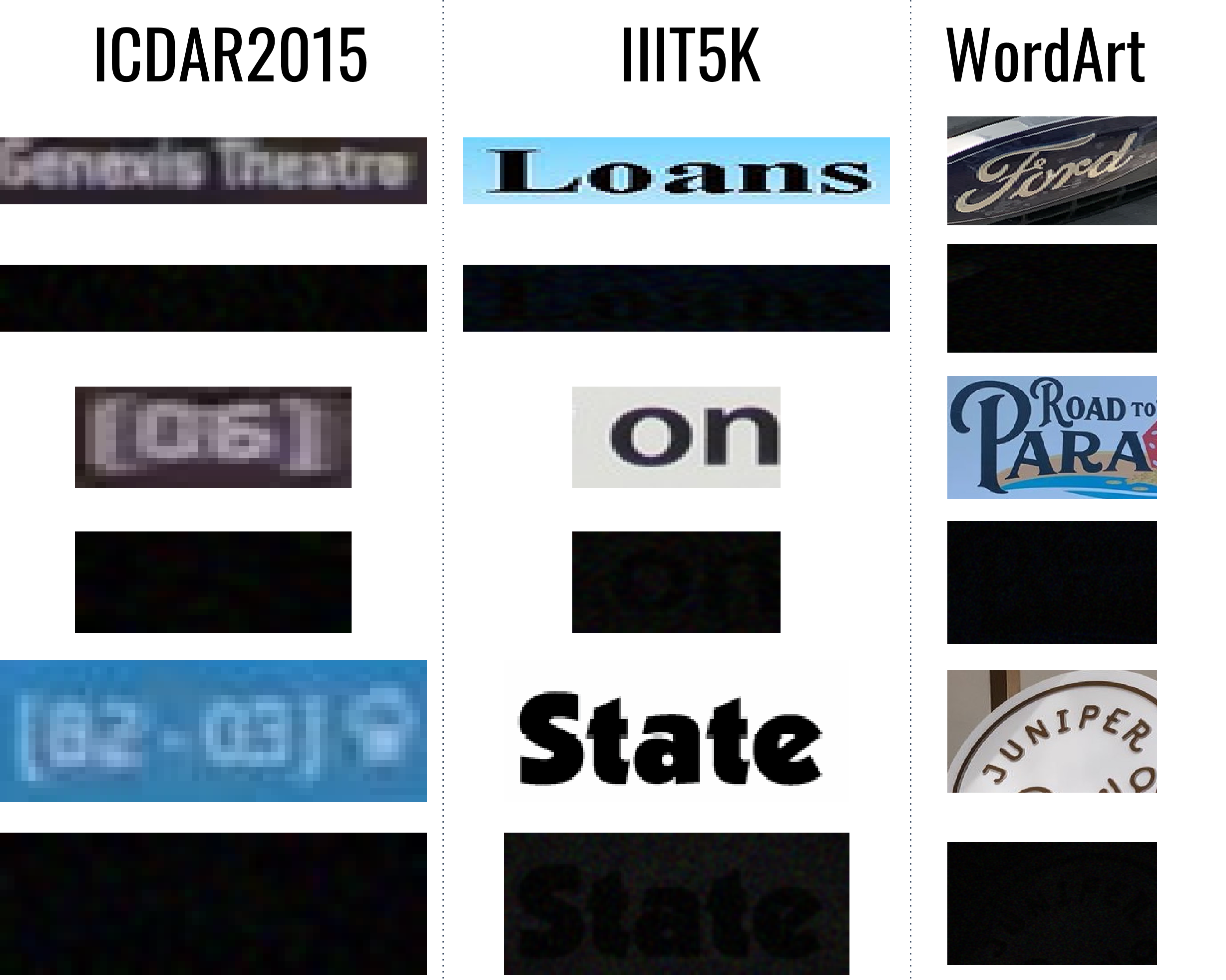}
   \caption{LSRT}
  \end{subfigure}

  \vspace{1mm}

  \begin{subfigure}{\linewidth}
    \centering
    \includegraphics[width=0.9\linewidth]{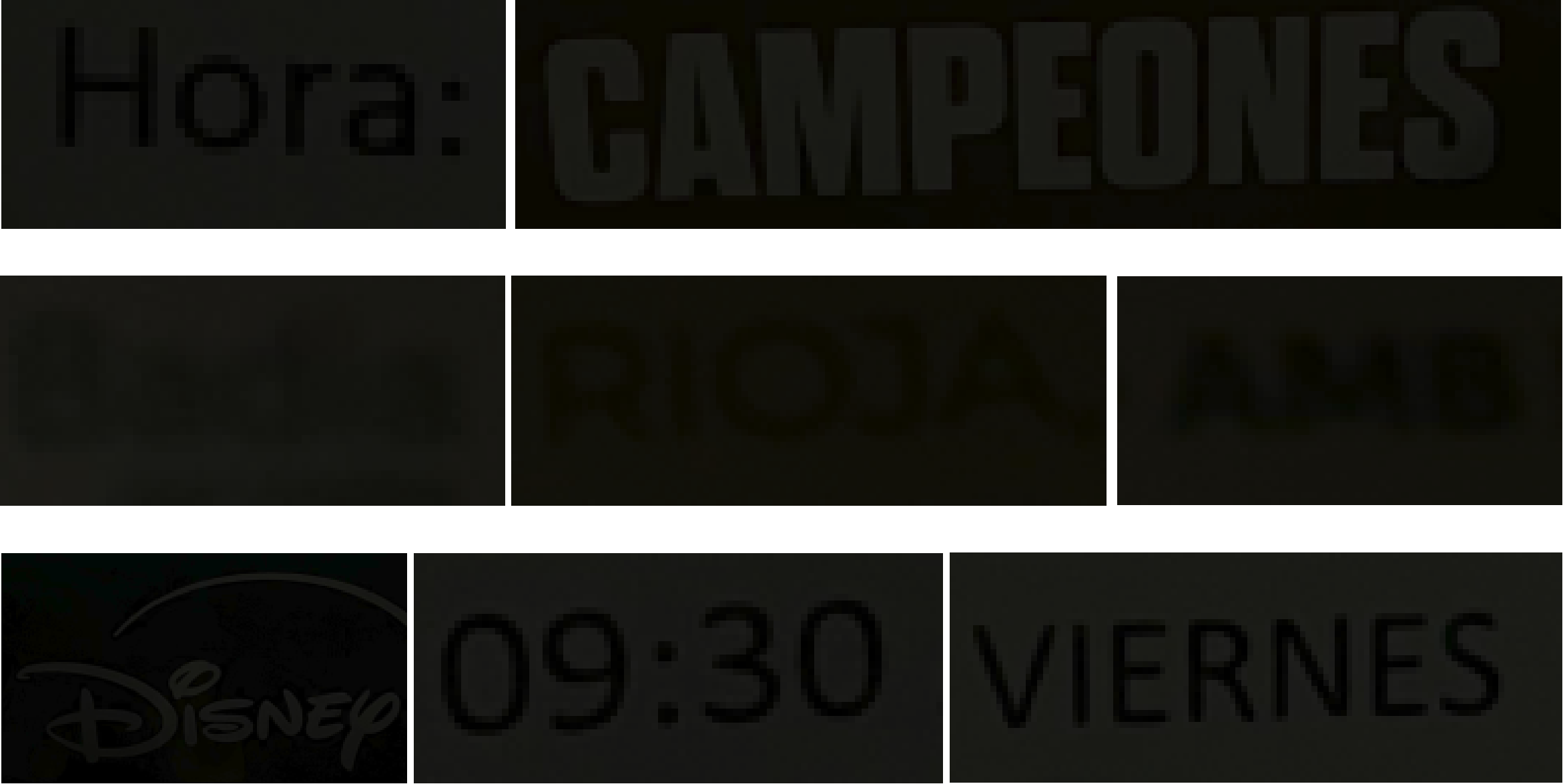}
    \caption{ESTR}
  \end{subfigure}

  \caption{Visualization of our proposed low-light scene text datasets \textbf{LSTR}  and \textbf{ESTR}.}
  \label{fig:data}
\end{minipage}%
\hfill
\begin{minipage}[t]{0.56\textwidth}
  \vspace{0pt}\centering

  \includegraphics[width=\linewidth]{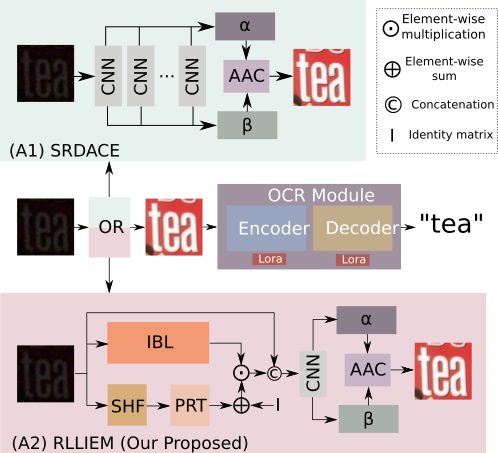}
  \caption{The main pipeline takes a low-light scene text image as input, enhances it using the LLIE module (A1. SRDACE~\cite{wen2023self} or A2. our proposed RLLIEM), and processes the enhanced image with a LoRA fine-tuned OCR module to predict the text "tea". }
  \label{fig:arch}
\end{minipage}%
\vspace{-4mm}
\end{figure*}

\subsection{Dataset Construction}
\label{sec:data}

We present two different datasets: one synthetically generated one LSTR and one real captures one ESTR.

\paragraph{\textbf{Synthetic dataset.}} we developed an algorithm to transform normally lit text recognition images into low-light images. We applied this algorithm to the ICDAR2015 \cite{peyrard2015icdar2015}, IIIT5K \cite{mishra2012scene}, and WordArt \cite{xie2022toward} datasets. The obtained low-light text recognition datasets were then used as a synthetic dataset for training. 

Our algorithm performs the following steps: Normalization, linear darkening, non-linear darkening, noise addition, vignetting effect, and (Optional) blurring effect. In what follows, we give more details on these different operations.

\textbf{\ding{172} Image normalization:} As the datasets considered are coming from completely uncontrolled scenarios, there are multiple illumination inconsistencies caused by strong lighting and shadows. To reduce these inconsistencies, we apply a normalization step by applying the Single Scale Retinex (SSR) algorithm \cite{Jobson97}. Thus, our first step is
\begin{equation}
    I_{norm}=SSR(I_{in})
\end{equation}

\textbf{\ding{173} Linear darkening:} Once the images have been normalized, we can then propose a darkening strategy for them. We start by performing a linear darkening that reduce the contrast of the image by a factor of $k$. Mathematically, we apply:
\begin{equation}
I_{\text{lindark}}(x,y)= k \cdot I_{norm}(x,y).
\end{equation}

\textbf{\ding{174} Non-linear darkening:} We also perform a non-linear transformation for increasing the darkening effect. 
\begin{equation}
I_{\text{dark}}(x,y)=I_{\text{lindark}}(x,y)^\gamma,
\end{equation}
where we use $\gamma=5$.

\textbf{\ding{175} Add Gaussian Noise:} 
Images under low-light conditions are affected by noise at a higher rate than those captured in well-lit conditions. Thus, we add white gaussian noise \( n(x,y)\sim\mathcal{N}(0,\sigma_n^2) \) where \( \sigma_n=\text{noise\_level}\times255 \) to the image following the Additive White Gaussian Noise (AWGN) assumption \cite{AWGN}:
\begin{equation}
I_{\text{noisy}}(x,y)=\mathrm{clip}\Bigl(I_{\text{dark}}(x,y)+n(x,y),0,255\Bigr).
\end{equation}

\textbf{\ding{176} Vignetting:} We apply a vignetting effect, as vignetting is more exacerbated in low-light capturing conditions. To do so, we apply a normalized 2D Gaussian mask \( V(x,y) \) (obtained via the product of two 1D Gaussian kernels): 
\begin{equation}
    I_{\text{vignette}}(x,y)=I_{\text{noisy}}(x,y)\cdot V(x,y).
\end{equation}

\textbf{\ding{177} Blurring (Optional):} Similarly to vignetting, blurring is more common in low-light capturing conditions, thus we also propose to simulate blurring in our synthetic dataset by convolving with a Gaussian kernel \( G_{\text{blur}} \):
\begin{equation}
  I_{\text{final}}(x,y)=\bigl(I_{\text{vignette}}*G_{\text{blur}}\bigr)(x,y).  
\end{equation}

\paragraph{\textbf{Real dataset.}} We also captured a real dataset, which we call ESTR. Our dataset consists of $60$ different scenes. They are nighttime images, covering various text instances with different fonts, languages (English and Spanish), and backgrounds under diverse low-light conditions.

\paragraph{\textbf{Scope of the Proposed Datasets.}} LSTR and ESTR are intended as compact benchmarks that facilitate research on low-light text recognition, rather than comprehensive datasets. LSTR provides a controlled synthetic environment for systematic evaluation, while ESTR offers a small but diverse real-world test set. Table \ref{tab:dataset}, shows the total number of images in each dataset and the original dataset they come from. We acknowledge that larger and more varied nighttime datasets are necessary for fully characterizing model generalization, and we plan to expand both datasets in subsequent work.


\begin{table}[t!]
\centering
\caption{Statistics of \textbf{LSTR}, synthesized from Lit datasets into low-light, and \textbf{ESTR}, a real-world nighttime street text dataset in English and Spanish.}
\begin{tabular}{ccccc}
\toprule
\textbf{Dataset}  & \textbf{Illum.}  & \textbf{Train}  & \textbf{Test}  & \textbf{Total}  \\
\midrule
ICDAR2015~\cite{peyrard2015icdar2015}  & Lit & 4,468   & 2,077  & 6,545   \\
IIIT5K~\cite{mishra2012scene}      & Lit  & 2,000   & 3,000  & 5,000   \\
WordArt~\cite{xie2022toward}     & Lit  & 4,805   & 1,511  & 6,316   \\
\textbf{LSTR (Our)} & \textbf{Dark} & \textbf{11,273}  & \textbf{6,588}  & \textbf{17,861}  \\
\hline
\textbf{ESTR (Our)}      & \textbf{Dark}    &   $-$  &   \textbf{60}  & \textbf{60}     \\
\bottomrule
\end{tabular}

\label{tab:dataset}
\end{table}

\subsection{SRDACE Module}

We first employ the existing lightweight low-light enhancement module SRDACE~\cite{wen2023self}. Its design defines, for each color channel $c \in \{r, g, b\}$, a differentiable and monotonic adaptive adjustment curve (AAC) that is used to enhance or suppress pixel intensities within the \([0,1]\) range. For each color channel \(c \in \{r, g, b\}\), the AAC is defined as:
\begin{equation}
\mathrm{AAC}^c(\alpha_i; \beta_i; I^c) 
= I^c + \alpha_i \otimes \frac{1}{\beta_i} \otimes S(\beta_i; I^c) \otimes I^c \otimes \bigl(\beta_i - I^c\bigr),
\end{equation}
where the function \(S\) is given by $S(\beta_i; I^c) = \mathrm{Sigmoid}\bigl(-\,I^c + \beta_i - 0.1\bigr)$,
\(\mathbf{I^c}\) is the normalized pixel intensity of the \(c\)-th color channel (red, green, or blue), taking values in \([0,1]\), \(\boldsymbol{\alpha_i}\) is a learnable scalar controlling the \emph{enhancement strength} with larger values of \(\alpha_i\) generally imply stronger adjustment (enhancement or suppression), \(\boldsymbol{\beta_i}\) is another learnable scalar that determines the \emph{enhancement threshold}, specifying the intensity region with the most pronounced enhancement or suppression, \(\mathbf{S(\beta_i; I^c)}\) is a sigmoid-based function that outputs a value between 0 and 1, distinguishing intensity ranges around \(\beta_i\) by producing larger values for stronger enhancement when \(-\,I^c + \beta_i - 0.1\) is positive, and smaller values for suppression or minimal enhancement otherwise.

\paragraph{OCR Module.} Throughout all experiments, the OCR component refers to TrOCR, a transformer-based encoder–decoder recognizer pretrained on large-scale text datasets. Unless otherwise specified, the term “OCR” in the methodology denotes this TrOCR backbone, which is either frozen, fully fine-tuned, or adapted via 
LoRA depending on the setting.

\subsection{Our Proposed RLLIE Module}

As shown in Figure~\ref{fig:arch}, our end-to-end framework couples a low-light enhancement front-end (SRDACE or our proposed RLLIE) with a LoRA-tuned TrOCR recognizer; in this subsection, we detail the proposed RLLIE module. Although low-light scene text recognition is fundamentally a 2D problem, the  degradations observed in real nighttime imagery, such as global attenuation, non-uniform illumination falloff, directional shadows, and specular distortions follow the same physical principles encoded in the rendering equation (Eq.~8). Generic LLIE models implicitly adjust brightness but do not provide explicit control over how illumination interacts with text regions, often leading to over-smoothing or inconsistent contrast restoration. 

To address this, we adopt lightweight neural approximations of IBL and PRT not to perform physically accurate 3D relighting, but to introduce a \emph{physics-guided re-lighting prior} that stabilizes enhancement under extreme underexposure. Unlike conventional relighting networks that require geometry, depth, or calibrated HDR lighting, our formulation treats IBL and PRT as learnable priors embedded into a 2D enhancement module. This design enables RLLIE to model global illumination effects in a way that is consistent across text regions, while remaining fully trainable end-to-end with OCR supervision.

\begin{figure}[t!]
\centering
\begin{minipage}[t]{0.49\linewidth}
  \vspace{0pt}\centering
  \includegraphics[width=0.8\linewidth]{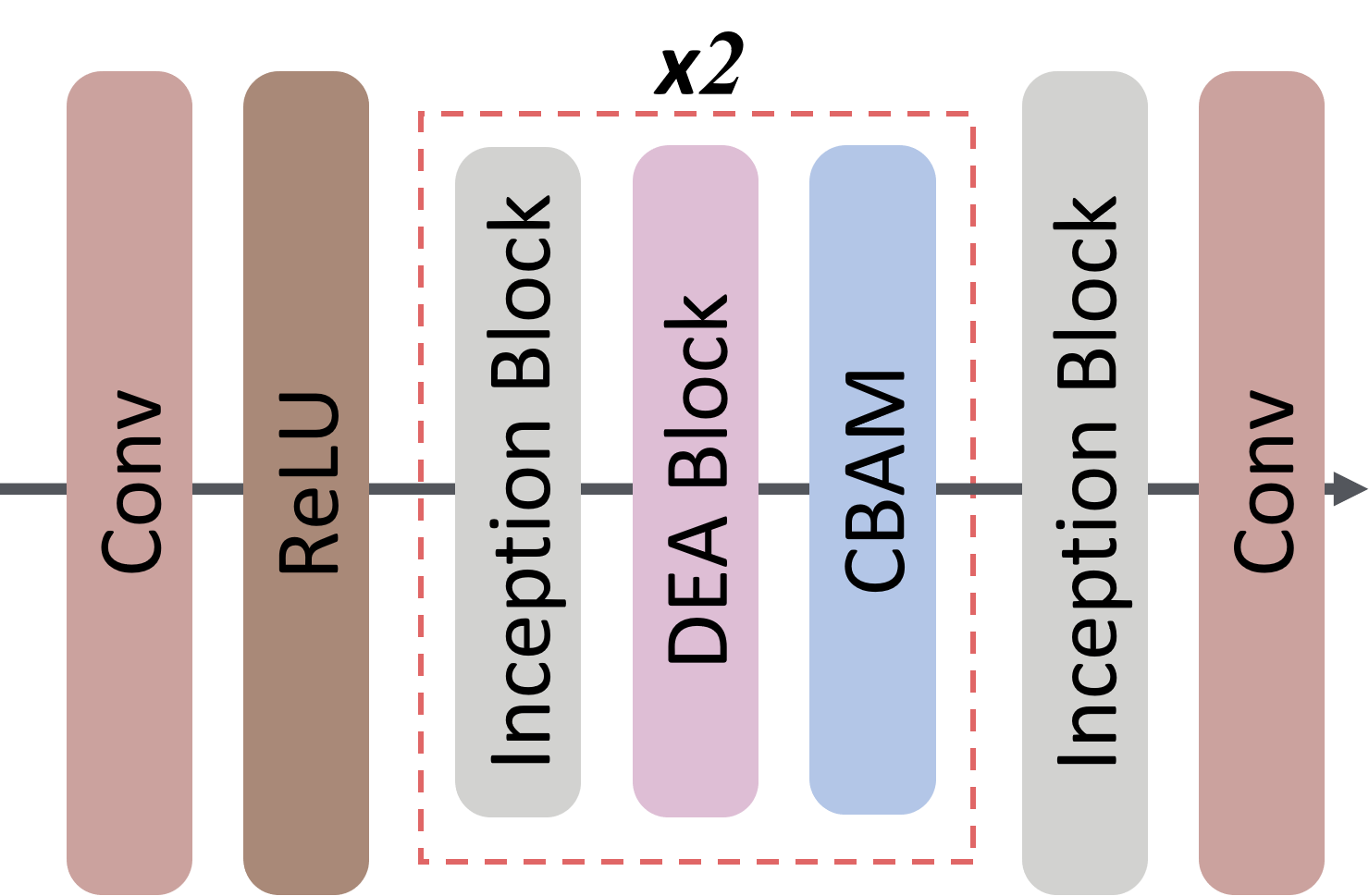}
  \caption{Our proposed IBL module.}
  \label{fig:ibl}
\end{minipage}%
\hfill
\begin{minipage}[t]{0.49\linewidth}
  \vspace{0pt}\centering
  \includegraphics[width=0.7\linewidth]{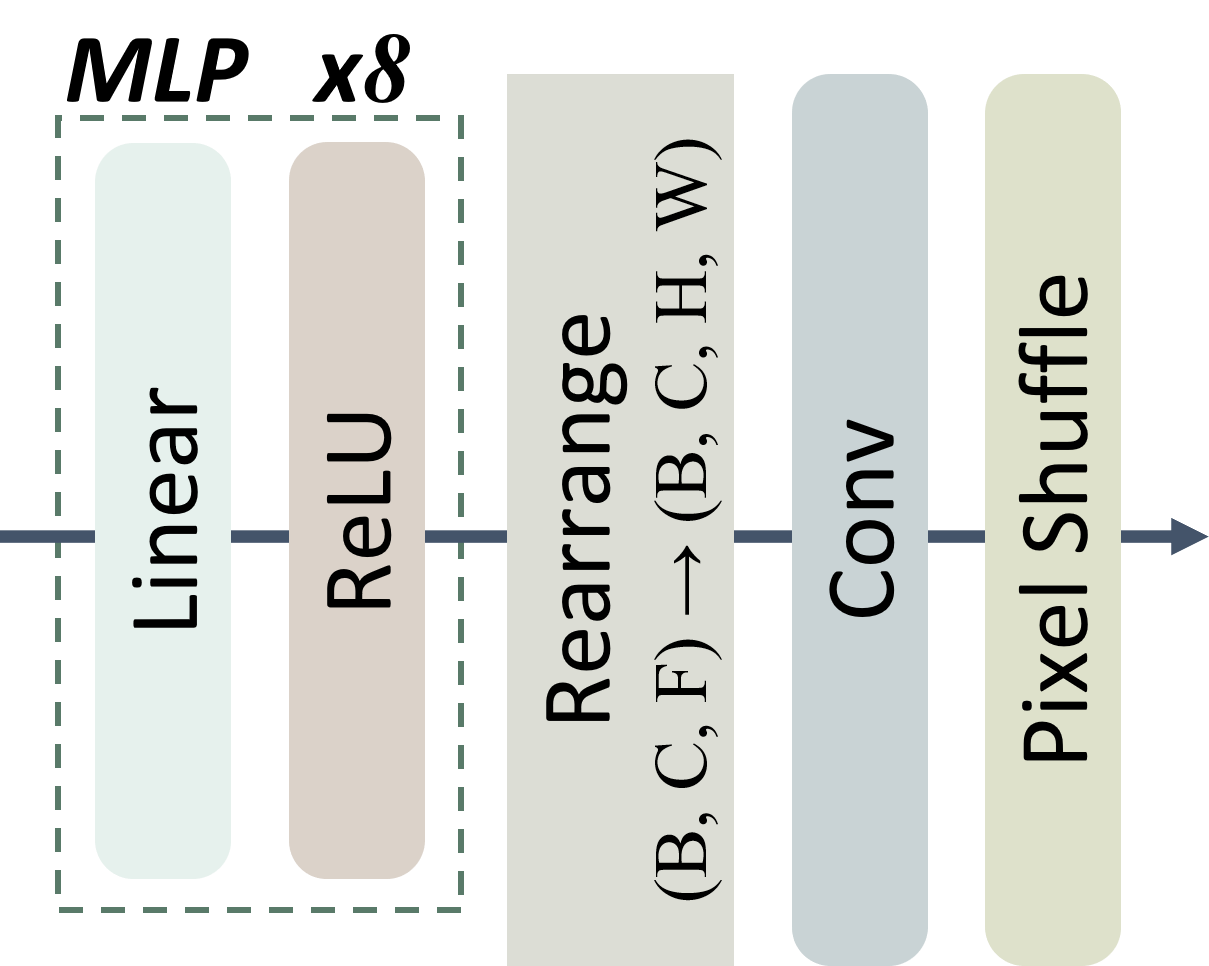}
  \caption{Our proposed PRT module.}
  \label{fig:prt}
\end{minipage}%
\vspace{-4mm}
\end{figure}

The rendering equation is an integral equation and serves as the theoretical basis for all global illumination methods, including ray tracing, path tracing, and radiosity. In computer graphics, photorealistic rendering aims to solve this equation. Two notable approaches are Image-Based Lighting (IBL) and Precomputed Radiance Transfer (PRT). However, both methods are computationally expensive. To address this, we simulate them by introducing our proposed neural network-based IBL (Figure~\ref{fig:ibl}) and PRT modules (Figure~\ref{fig:prt}). 


First, we introduce the \textbf{Unified Rendering Equation} (URE) for global illumination:
\begin{equation} \label{eq:rendering}
  L_o(\mathbf{x}, \omega_o) = \int_{\Omega} f_r(\mathbf{x}, \omega_i, \omega_o)\, L_i(\mathbf{x}, \omega_i)\, (\omega_i \cdot \mathbf{n})\, d\omega_i,
\end{equation}
where $L_o$ and $L_i$ denote the outgoing and incoming radiance, $f_r$ is the BRDF, and $(\omega_i \cdot \mathbf{n})$ is the cosine term. $\Omega$ is the incident-direction domain (typically the hemisphere, or the full sphere under environment lighting). To approximate Eq.~\ref{eq:rendering} efficiently, we build upon two representative paradigms: \textbf{IBL} and \textbf{PRT}, and design neural modules to emulate this rendering process.


\textbf{\ding{172} IBL} uses an HDR environment map as illumination, i.e., 
$L_i(\mathbf{x},\omega_i)=L_{\text{env}}(\omega_i)$.

\textbf{Diffuse:} For Lambertian surfaces, URE's outgoing diffuse radiance can be thus written as
\begin{equation} \label{eq:diffuse}
  L_{o,\text{diffuse}}(\mathbf{x}) = \frac{\rho}{\pi}\int_{\Omega} L_{\text{env}}(\omega_i)\,\max(0,\omega_i\!\cdot\!\mathbf{n})\,d\omega_i,
\end{equation}
which corresponds to irradiance from the environment.

\textbf{Specular:} For glossy/specular materials, we adopt a microfacet-based formulation and evaluate the
environment integral using standard IBL approximations (e.g., prefiltered environment maps and a view-dependent lookup),
instead of explicitly computing the full BRDF integral per pixel.

As illustrated in Figure~\ref{fig:ibl}, we design a lightweight neural IBL module as a \emph{physics-guided prior} within RLLIE to approximate environment-driven illumination effects in Eq.~\ref{eq:rendering}.  Given an input HDR environment map, a $3\times3$ convolution with ReLU extracts low-level lighting features, which are then processed by a cascaded Inception--DEA--CBAM submodule.  Specifically, the Inception Block~\cite{szegedy2015going} captures multi-scale lighting structures via multi-branch convolutions; the DEA Block~\cite{chen2024dea} adaptively reallocates channel capacity to highlight salient illumination components; and CBAM~\cite{woo2018cbam} refines the representation through channel- and spatial-wise attention.  Overall, this cascade learns a multi-scale decomposition and adaptive aggregation of environment illumination, jointly modeling the diffuse irradiance response and the glossy

\textbf{\ding{173} PRT} precomputes light transport effects, such as shadows and indirect illumination, for static scenes.


\textbf{Light Transport:} At a surface point $\mathbf{x}$, we define the transport weight for an incoming direction $\omega$ as
\begin{equation} \label{eq:transport}
  T(\mathbf{x}, \omega) = V(\mathbf{x}, \omega) \, \max(0, \mathbf{n}(\mathbf{x}) \cdot \omega),
\end{equation}
where $V(\mathbf{x},\omega)$ denotes visibility (1 if unoccluded, 0 otherwise). Under environment illumination, the outgoing radiance is
\begin{equation} \label{eq:prt}
  L_o(\mathbf{x}) = \int_{S^2} T(\mathbf{x}, \omega) \, L_{\text{env}}(\omega) \, d\omega.
\end{equation}

To obtain a compact and efficient angular representation, we parameterize $T(\mathbf{x},\omega)$ using spherical harmonics (SH). The predicted SH coefficients at $\mathbf{x}$ can be interpreted as directional attention weights over a fixed set of angular bases, enabling efficient encoding of light transport. As shown in Figure~\ref{fig:prt}, we propose a neural PRT module that maps geometric cues at $\mathbf{x}$ (e.g., position, normal, and visibility encoding) to the corresponding SH coefficient vector $\{t_{l,m}(\mathbf{x})\}$. The architecture consists of an 8-layer MLP (Linear + ReLU) to model the nonlinear mapping from input features to transport coefficients, followed by a feature rearrangement stage that reshapes the output from $(B,C,F)$ to $(B,C,H,W)$ for convolutional processing. A convolutional context encoder is then applied to capture local spatial correlations and improve coefficient consistency, and a pixel-shuffle block upsamples the features to produce a dense SH coefficient map.

We emphasize that RLLIE does not seek to propose a new rendering formulation. Instead, it draws inspiration from classical illumination transport and instantiates it as learnable, task-driven priors for low-light text recognition. In contrast to neural relighting approaches that typically assume explicit scene geometry/depth and optimize for photorealistic appearance, our IBL/PRT modules operate entirely in the 2D feature space and are trained end-to-end with OCR supervision. This design encourages enhancement that preserves text-relevant structures (e.g., fine strokes and sharp boundaries) critical for recognition, positioning RLLIE as a task-oriented adaptation of illumination modeling rather than a general-purpose relighting system.

\subsection{Loss Functions}

Our end-to-end pipeline is optimized using two components: an \textbf{LLIE loss} for enhancement and an \textbf{OCR loss} for recognition. We detail them in the below.

\paragraph{\textbf{\textit{LLIE Loss}}.}
For LLIE supervision, we integrate the widely used low-light enhancement loss $\mathcal{L}_{\mathrm{LLE}}$~\cite{wen2023self}, which comprises four components: Localized Color Consistency (LCC), Global Chromatic Fidelity (GCF), Brightness Preservation (BP), and Curve Regularization (CR), with our proposed Edge content loss $\mathcal{L}_{\mathrm{EC}}$. 


\paragraph{Edge Content Loss.}
To encourage preservation of text-relevant structures, we define an Edge Content Loss as the mean squared error between Sobel edge maps of the low-light input $I_{\mathrm{low}}$ and the enhanced output $I_{\mathrm{out}}$:
\begin{equation}
\mathcal{L}_{\mathrm{EC}} = \frac{1}{N_p} \sum_{i=1}^{N_p} \left( S(I_{\mathrm{out}})_i - S(I_{\mathrm{low}})_i \right)^2,
\end{equation}
where $N_p$ is the number of pixels and $S(\cdot)_i$ denotes the Sobel operator response at pixel $i$. $\varphi$ controls the contribution of $\mathcal{L}_{\mathrm{EC}}$ and is set to balance the relative magnitudes of different terms.

\paragraph{Overall LLIE Loss.}
The full LLIE objective is a weighted sum:
\begin{equation}
\mathcal{L}_{\mathrm{LLIE}} = \mathcal{L}_{\mathrm{LLE}} + \varphi \mathcal{L}_{\mathrm{EC}},
\end{equation}
where $\varphi$ denotes the weighting factor and is set to $1\times10^5$. 
\paragraph{\textbf{\textit{OCR Loss}}.}
We use the standard token-level cross-entropy loss on the decoder logits $z \in \mathbb{R}^{N_t \times V}$:
\begin{equation}
\mathcal{L}_{\mathrm{OCR}} = -\frac{1}{N_t} \sum_{i=1}^{N_t} \log \left( \frac{\exp(z_{i,y_i})}{\sum_{j=1}^{V} \exp(z_{i,j})} \right),
\end{equation}
where $y_i$ is the ground-truth token index for token $i$, $V$ is the vocabulary size, and $N_t$ is the total number of tokens.

\paragraph{\textbf{\textit{Total Loss}}.}
The total training objective is:
\begin{equation}
\mathcal{L}_{\mathrm{total}} = \mathcal{L}_{\mathrm{LLIE}} + \omega \mathcal{L}_{\mathrm{OCR}},
\end{equation}
where $\omega$ denotes the weighting factor and is set to $100$. 

The different weights were selected on a small validation sweep and kept constant across experiments.

\section{Experiment}

\subsection{Experimental Analysis}

As reported in Table \ref{tab:recognition_performance}, we evaluate three TrOCR adaptation strategies: frozen backbone (“*”), full fine tuning and LoRA, using CER as the metric. Both frozen and fully fine‑tuned models yield 53.16\% CER, showing that naively updating all weights on scarce LSTR data offers no benefit. In contrast, LoRA tuning cuts CER to 52.84\% (a statistically significant 0.32\% drop versus ±0.10\% variance), demonstrating that parameter‑efficient adaptation outperforms zero‑shot inference and full fine‑tuning when in‑domain data are scarce or a substantial domain gap exists. The identical CERs for frozen and fully fine‑tuned models suggest that LSTR’s training set may be too small or too noisy for stable end‑to‑end optimization. LoRA’s advantage stems from updating only a low‑rank subspace, thereby preserving most pretrained knowledge while capturing LSTR’s specific characteristics; its smaller footprint and faster convergence also make it preferable in production, where even a 0.32\% CER reduction can meaningfully reduce manual correction effort.

Table \ref{tab:recognition_comparison} compares four low‑light image enhancement and OCR pipelines on LSTR. Pretrained TrOCR alone scores 53.16\% CER, reflecting its failure on low‑contrast layouts. Prepending frozen SRDACE degrades performance to 53.57\% because generic enhancement can distort strokes for a static OCR. Joint fine‑tuning of SRDACE with TrOCR reduces CER to 50.02\%, demonstrating that enhancement must be optimized for the OCR loss to boost text‑region contrast without introducing artifacts. Incorporating RLLIE—an end‑to‑end re‑rendering based enhancer that uses image‑based lighting, precomputed radiance transfer and text‑aware gating—yields 50.43\% CER, nearly matching SRDACE’s best result while better preserving stroke fidelity under extreme underexposure. The minor gap arises from RLLIE’s global illumination focus, which can slightly vary boundary thickness, but its physics‑grounded re‑lighting losses align reliably with TrOCR objectives. These results confirm that task‑aware enhancement, whether data‑driven or physics‑inspired, is essential to bridge the low‑light domain gap for Transformer‑based OCR.

\begin{table}[t!]
\centering

\begin{minipage}[t]{0.48\linewidth}
\centering
\captionof{table}{TrOCR performance on \textbf{LSTR} using direct evaluation, fine-tuning, and LoRA-tuning.}
\label{tab:recognition_performance}
\resizebox{\linewidth}{!}{
\begin{tabular}{cccc}
\toprule
\textbf{Method}   & \textbf{Training} & \textbf{Dataset} & \textbf{CER$\downarrow$}\\
\midrule
TrOCR & \raisebox{-0.5ex}{*} & LSTR & 53.16\\
TrOCR & fine-tune            & LSTR & 53.16 \\
TrOCR & LoRA-tune            & LSTR & \textbf{52.84}\\
\bottomrule
\end{tabular}
}
\end{minipage}
\hfill
\begin{minipage}[t]{0.48\linewidth}
\centering
\captionof{table}{Combinations of LLIE and OCR modules with * indicating frozen weights.}
\label{tab:recognition_comparison}
\resizebox{\linewidth}{!}{
\begin{tabular}{llcc}
\toprule
\textbf{LLIE} & \textbf{OCR} & \textbf{Dataset} & \textbf{CER$\downarrow$}\\
\midrule
$-$      & TrOCR* & LSTR & 53.16 \\
SRDACE*  & TrOCR* & LSTR & 53.57\\
SRDACE   & TrOCR  & LSTR & \textbf{50.02}\\
RLLIE    & TrOCR  & LSTR & 50.43\\
\bottomrule
\end{tabular}
}
\end{minipage}

\end{table}

\subsection{Evaluation of Real-World Nighttime Scene Text Recognition}

\begin{figure}[t!]
  \centering
  \begin{minipage}[t]{0.55\linewidth}
    \vspace{0pt}
    \centering
    \includegraphics[width=\linewidth]{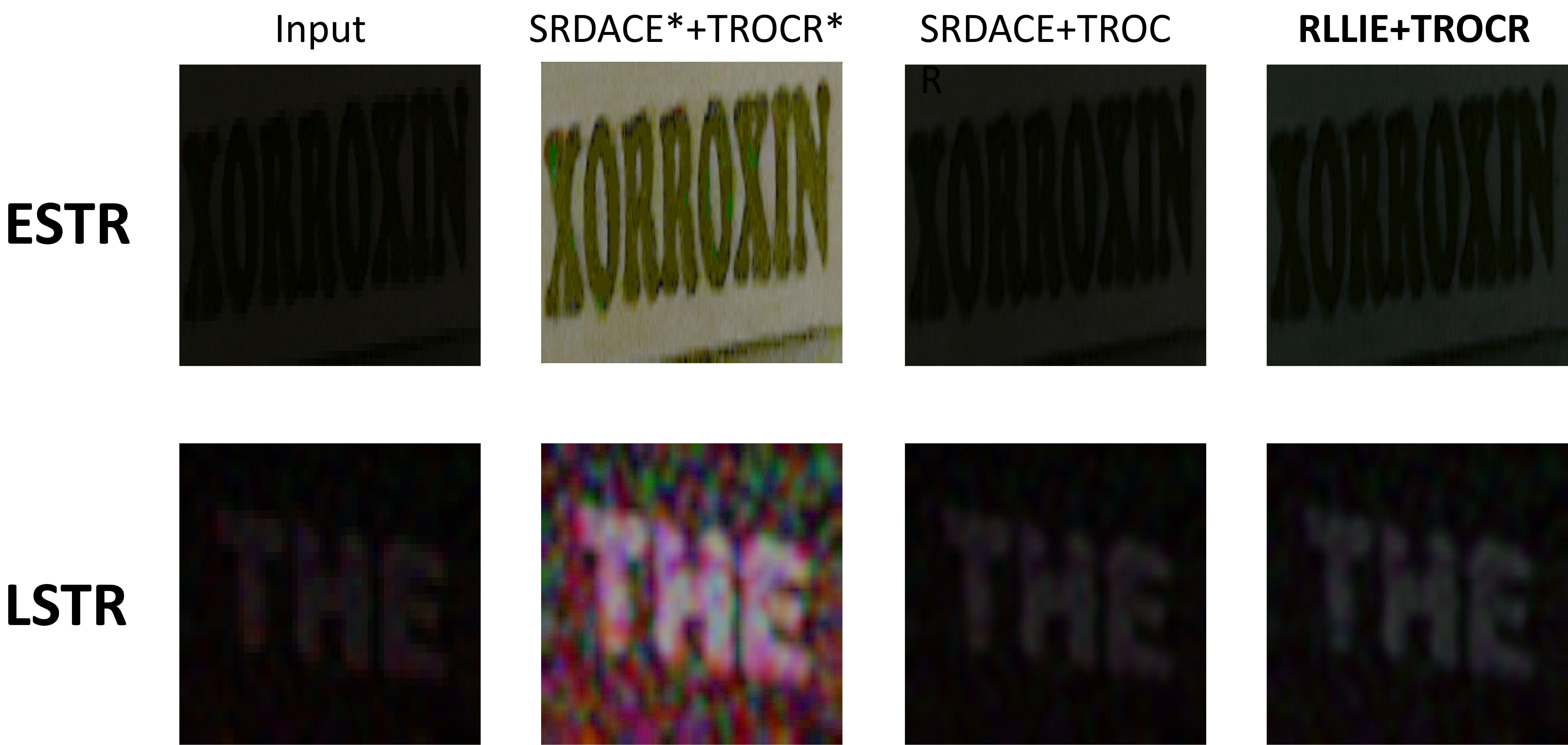}
    \caption{Visualization of enhanced scene text images from the \textbf{LSTR} and \textbf{ESTR} datasets.}
    \label{fig:reault}
  \end{minipage}
  \hfill
  \begin{minipage}[t]{0.42\linewidth}
    \vspace{0pt}
    \centering
    \captionof{table}{Direct evaluation on \textbf{ESTR} dataset.}
    \label{tab:zero_shot_real}
    \vspace{2mm}
    \begin{tabular}{l l c}
      \toprule
      \textbf{Method} & \textbf{Dataset} & \textbf{CER$\downarrow$}\\
      \midrule
      TrOCR* & ESTR & 25.26\\
      SRDACE+TrOCR & ESTR & 14.43\\
      SRDACE*+TrOCR* & ESTR & 23.20\\
      RLLIE & ESTR & \textbf{13.66}\\
      \bottomrule
    \end{tabular}
  \end{minipage}

  \vspace{-3mm}
\end{figure}


We evaluated our model directly on the real‐world low‐light dataset ESTR without additional training (Table \ref{tab:zero_shot_real}). TrOCR* yields a CER of 25.26\%, demonstrating its inability to handle ESTR’s low contrast, uneven illumination, and varied handwriting/font styles. Introducing a frozen SRDACE front‐end reduces CER marginally to 23.20\%, as a fixed correction cannot fully address content‐specific distortions nor allow the recognizer to adapt. Jointly fine‐tuning SRDACE on ESTR and TrOCR’s encoder–decoder shrink CER dramatically to 14.43\%, confirming that end‐to‐end optimization of enhancement and recognition is essential under severe low‐light conditions. Our RLLIE method, which re‐renders images via learned lighting from environment maps and text‐region attention without backpropagating through TrOCR, achieves a CER of 13.66\%. By respecting global geometry and preserving stroke details, RLLIE generalizes more robustly than generic LLIE modules, closing most of the domain gap in a zero‐shot setting.

\paragraph{Discussion on Synthetic–Real Performance Gap.} A seemingly counter-intuitive observation is that all models perform  significantly worse on the synthetic LSTR dataset than on the real ESTR data. This discrepancy arises from two factors. First, LSTR intentionally follows a \emph{stress-test} design: the non-linear suppression, strong AWGN, and vignetting jointly produce contrast levels more severe than those typically found in real nighttime street scenes. As a result, LSTR is considerably more challenging than ESTR. Second, ESTR predominantly contains larger and more structured text instances (e.g., street signs), whereas LSTR includes many small, fine-grained crops from datasets such as WordArt, making recognition inherently harder. We acknowledge this limitation and plan to refine the synthetic pipeline and expand ESTR in future work.



\subsection{Impact of Image Brightness on OCR Performance}

\paragraph{Clarification on Brightness and Recognition.} Our claim that “brightness optimization alone may degrade recognition” does not  suggest that any increase in illumination is inherently harmful to OCR. As  shown in Table~4, applying a frozen SRDACE module yields a small improvement in  CER on ESTR. The key observation, however, is that \emph{aggressive  brightness-maximizing enhancement tends to over-correct}, introducing haloing,  edge diffusion, and color shifts that disrupt the integrity of text strokes.  Joint training mitigates this issue by aligning the enhancement process with the  OCR gradients, preventing excessive amplification of illumination and enabling  the model to restore visibility while preserving character structure. In this  sense, brightness is beneficial only when modulated under task-aware  supervision.

The RLLIE TrOCR framework leverages inherent lighting cues to re-render  low-light images, enhancing local detail and refining texture. This produces  slightly cooler-toned outputs that facilitate higher-quality feature extraction  by TrOCR. End-to-end joint optimization further preserves edges, contours, and  chromatic consistency, resulting in improved recognition accuracy. As shown in  Figure~\ref{fig:reault}, RLLIE-enhanced images exhibit sharper boundaries and  stronger contrast, enabling more reliable OCR.

In comparison, SRDACE TrOCR provides only modest benefits: although it improves  overall visibility, it fails to reconstruct fine texture, leading to blurred  strokes (Figure~\ref{fig:reault}) and hindering TrOCR's ability to extract clean  character outlines. This underscores that generic low-light enhancement is  insufficient for recognition-sensitive tasks.

Finally, the frozen SRDACE* TrOCR* configuration passes enhanced images into an  unchanged OCR model, resulting in mismatched objectives between enhancement and  recognition. This supports \textbf{RQ1}: increasing brightness alone—even when  visually pleasing to humans—does not guarantee improved OCR performance and may  sometimes degrade it.

Although one could further decompose RLLIE into its constituent components such  as IBL, PRT, or individual loss terms, the primary aim of this study is to  evaluate the \emph{integrated} enhancement–recognition pipeline under realistic  low-light conditions. The empirical results across both LSTR and ESTR reveal two  consistent and robust findings: (1) task-aware enhancement substantially  outperforms generic or frozen LLIE approaches, and (2) incorporating  physics-guided priors stabilizes illumination correction and improves the  preservation of text-relevant structures. These trends hold across datasets and  settings, indicating that the present system-level evaluation is sufficient to  support the central claims of the paper. Additional ablations, while potentially  useful for future investigations, are not necessary to validate the effectiveness  of the proposed integrated formulation.

\section{Conclusion}

We introduce a challenging low‑light scene text recognition benchmark and systematically evaluate fusion strategies between LLIE and OCR, revealing that nighttime artifacts, low contrast, noise, and blur, weaken standalone methods and motivate a unified baseline. We propose RLLIE, a re‑rendering LLIE module that significantly boosts OCR accuracy on both synthetic and real datasets. Future work will expand our dataset with diverse nighttime imagery and explore feature‑level alignment and fusion of LLIE and OCR for robust text recognition under extreme low‑light conditions.

\section{Acknowledgements}

This work has been partially supported by the predoctoral program AGAUR-FI ajuts (2026 FI-3 00470) Joan Oró, which are backed by the Secretariat of Universities and Research of the Department of Research and Universities of the Generalitat of Catalonia, as well as the European Social Plus Fund; the Beatriu de Pinós del Departament de Recerca i Universitats de la Generalitat de Catalunya (2022 BP 00256); the Grant PID2021-128178OB-I00, PID2023-146426NB-100, PID2024-162555OB-I00 funded by MICIU/AEI/10.13039/501100011033, ERDF, EU ``A way of making Europe'' and by the Generalitat de Catalunya CERCA Program; the European Social Plus Fund, European Lighthouse on Safe and Secure AI (ELSA) from the European Union’s Horizon Europe programme under grant agreement No 101070617. JVC also acknowledges the 2025 Leonardo Grant for Scientific Research and Cultural Creation from the BBVA Foundation. The BBVA Foundation accepts no responsibility for the opinions, statements and contents included in the project and/or the results thereof, which are entirely the responsibility of the authors.

%
%
%
\bibliographystyle{splncs04}
%




\bibliography{egbib2}

\end{document}